\documentclass[aos,compress,preprint]{imsart}
\setattribute{journal}{name}{}
\usepackage[round]{natbib}
\usepackage{graphicx}
\usepackage{amsmath,amsthm}
\usepackage{amsfonts,amssymb,mathrsfs}
\usepackage{bbm}
\usepackage{subfig}
\usepackage{mathtools}
\usepackage{enumerate}
\usepackage{fullpage}
\usepackage{bm}
\usepackage{url}
\usepackage[colorlinks=true,pagebackref,linkcolor=magenta]{hyperref}
\usepackage{parskip}
\newtheorem{theorem}{Theorem}

\newtheorem{corollary}{Corollary}

\theoremstyle{definition}
\newtheorem{definition}{Definition}
\newtheorem*{remark}{Remark}
\numberwithin{equation}{section}
\mathchardef\ordinarycolon\mathcode`\:
\mathcode`\:=\string"8000
\begingroup \catcode`\:=\active
  \gdef:{\mathrel{\mathop\ordinarycolon}}
\endgroup

\renewcommand{\tilde}{\widetilde}
\renewcommand{\Pr}{\mathbb{P}}

\renewcommand{\Pr}{\mathbb{P}}

\makeatletter
\def\thm@space@setup{%
 \thm@preskip=\parskip \thm@postskip=0pt
}
\makeatother

\begin{document}
\begin{frontmatter}
\title{The eigenvalues of stochastic blockmodel graphs}
\begin{aug} 
\author{\fnms{Minh}
\snm{Tang}\ead[label=e1]{minh@jhu.edu}}
 \runauthor{Minh Tang}
 \affiliation{Johns Hopkins University}
\address{Department of Applied Mathematics and Statistics,\\Johns
  Hopkins University,\\3400 N. Charles St,\\Baltimore, MD 21218, USA.\\
\printead{e1}}
\end{aug}
\begin{abstract} We derive the joint limiting distribution for the largest
eigenvalues of the adjacency matrix for stochastic blockmodel graphs
when the number of vertices tends to infinity. We show that, in
the limit, these eigenvalues are jointly multivariate normal with
bounded covariances. Our result extends the classical result of
F\"{u}redi and Koml\'{o}s on the fluctuation of the largest eigenvalue
for Erd\H{o}s-R\'{e}nyi graphs.
\end{abstract}
\end{frontmatter}

\section{Introduction} The systematic study of eigenvalues of random matrices 
dates back to the seminal work of \cite{Wigner} on the semicircle law
for Wigner ensembles of symmetric or Hermitean matrices. A random $n
\times n$ symmetric matrix $\mathbf{A} = (a_{ij})_{i,j=1}^{n}$ is said
to be a Wigner matrix if, for $i \leq j$, the entries $a_{ij}$ are independent
mean zero random variables with variance $\sigma^2_{ij} = 1$ for $i <
j$ and $\sigma_{ii}^2 = \sigma^2 > 0$. Many important and beautiful
results are known for the spectral properties of these matrices, such
as universality of the semi-circle law for bulk eigenvalues
\citep{Erdos2,Tao2}, universality of the Tracy-Widom distribution for
the largest eigenvalue \citep{Soshnikov}, universality properties of
the eigenvectors \citep{tao2012random,knowles}, and eigenvalue and
eigenvector delocalization \citep{Erdos3}.

In contrast, much less is known about the spectral properties of random symmetric matrices
$\mathbf{A} = (a_{ij})_{i,j=1}^{n}$ where the entries $a_{ij}$ are independent
but not necessarily mean zero random variables with possibly heterogeneous variances.  
Such random matrices arise naturally in many
settings, with the most popular example being perhaps the adjacency
matrices of (inhomogeneous) independent edge random graphs. 
In the case when $\mathbf{A}$ is the
adjacency matrix for an Erd\H{o}s-R\'{e}nyi graph where the edges are
i.i.d. Bernoulli random variables, \cite{Arnold} and \cite{ding10}
show that the empirical distribution of the eigenvalues of
$\mathbf{A}$ also converges to a semi-circle law. Meanwhile, the
following result of \citet{furedi1981eigenvalues} shows that the
largest eigenvalue of $\mathbf{A}$ is normally distributed when $\mathbb{E}[a_{ij}] = \mu$ and
$\mathrm{Var}[a_{ij}] = \sigma^2$ for $i < j$.
\begin{theorem}[\cite{furedi1981eigenvalues}]
\label{thm:furedi}
  Let $\mathbf{A} = (a_{ij})$ be an $n \times n$ symmetric matrix where
  the $a_{ij}$ are independent (not necessarily identically
  distributed) random variables uniformly bounded in magnitude by a constant
  $C$. Assume that for $i > j$, the $a_{ij}$ have a common expectation
  $\mu > 0$ and variance $\sigma^2$. Furthermore, assume that
  $\mathbb{E}[a_{ii}] = v$ for all $i$. Then the distribution of
  $\lambda_1(\mathbf{A})$, the largest eigenvalue of $\mathbf{A}$, can
  be approximated in order $n^{-1/2}$ by a normal distribution with
  mean $(n - 1) \mu + v + \sigma^2/\mu$ and variance $2
  \sigma^2$, i.e.,
  \begin{equation}
    \label{eq:1}
    \lambda_1(\mathbf{A}) - (n-1) \mu - v
    \overset{\mathrm{d}}{\longrightarrow} \mathcal{N}\bigl( \tfrac{\sigma^2}{\mu}, 2
    \sigma^2 \bigr)
  \end{equation} as $n \rightarrow \infty$. 
  Furthermore, with probability tending to $1$, 
  \begin{equation}
    \label{eq:2}
    \max_{i \geq 2} |\lambda_i(\mathbf{A})| < 2 \sigma \sqrt{n} +
    O(n^{1/3} \log n).
  \end{equation}
\end{theorem} In the case when $\mathbf{A}$ is the adjacency matrix
of an Erd\H{o}s-R\'{e}nyi graph with edge probability $p$,
Theorem~\ref{thm:furedi} yields
$$\lambda_1(\mathbf{A}) - np \overset{\mathrm{d}}{\longrightarrow} \mathcal{N}(1-p, 2p(1-p))$$
as $n \rightarrow \infty$. 

A natural generalization of Erd\H{o}s-R\'{e}nyi random graphs is the
notion of stochastic blockmodel graphs \citep{holland} where, given an
integer $K \geq 1$, the $a_{ij}$ for $i \leq j$ are independent
Bernoulli random variables with $\mathbb{E}[a_{ij}] \in \mathcal{S}$
for some set $\mathcal{S}$ of cardinality $K(K+1)/2$. More
specifically, we have the following definition.
\begin{definition}
  \label{def:SBM}
  Let $K \geq 1$ be a positive integer and let $\bm{\pi} = (\pi_1,
  \pi_2, \dots, \pi_K)$ be a non-negative vector in $\mathbb{R}^{K}$
  with $\sum_{k} \pi_k = 1$. Let $\mathbf{B} \in [0,1]^{K \times K}$
  be symmetric. We say that $(\bm{\tau}, \mathbf{A}) \sim
  \mathrm{SBM}(\bm{\pi}, \mathbf{B})$ if the following holds. First,
  $\bm{\tau} = (\tau_1, \dots, \tau_n)$ and the $\tau_i$ are
  i.i.d. random variables with $\Pr[\tau_i = k] = \pi_k$. Then
  $\mathbf{A} \in \{0,1\}^{n \times n}$ is a symmetric matrix such
  that, conditioned on $\bm{\tau}$, for all $i \geq j$ the $a_{ij}$
  are independent Bernoulli random variables with $\mathbb{E}[a_{ij}] = \mathbf{B}_{\tau_i, \tau_j}$.
\end{definition} 
The stochastic blockmodel is among the most
popular generative models for random graphs with community structure;
the nodes of such graphs are partitioned into blocks or communities,
and the probability of connection between any two nodes is a function
of their block assignment. The adjacency matrix $\mathbf{A}$ of a
stochastic blockmodel graph can be viewed as $\mathbf{A} =
\mathbb{E}[\mathbf{A}] + (\mathbf{A} -\mathbb{E}[\mathbf{A}])$ where
$\mathbb{E}[\mathbf{A}]$ is a low-rank deterministic matrix and
$(\mathbf{A} -\mathbb{E}[\mathbf{A}])$ is a generalized Wigner
matrix whose elements are independent mean zero random variables with
heteregoneous variances. We emphasize that our assumptions on 
$\mathbf{A} - \mathbf{E}[\mathbf{A}]$ distinguish us from existing results in
the literature. For example, \cite{peche,knowles_yin,bordenave,pizzo} consider finite rank
additive perturbations of the random matrix $\mathbf{X}$ given by
$\tilde{\mathbf{X}} = \mathbf{X} + \mathbf{P}$ under the assumption that
$\mathbf{X}$ is either a Wigner matrix or is sampled from the
Gaussian unitary ensembles. Meanwhile, in \cite{benaych-georges}, the authors assume that $\mathbf{X}$
or $\mathbf{P}$ is orthogonally invariant; a
symmetric random matrix $\mathbf{H}$ is orthogonally invariant if its distribution
is invariant under similarity transformations $\mathbf{H} \mapsto
\mathbf{W}^{-1} \mathbf{H} \mathbf{W}$ whenever $\mathbf{W}$ is
an orthogonal matrix. Finally, in \cite{O_Rourke}, the entries of $\mathbf{X}$
are assumed to be from an elliptical family of distributions, i.e.,
the collection $\{(X_{ij}, X_{ji})\}$ for $i < j$ are i.i.d. according
to some random variable $(\xi_1, \xi_2)$ with $\mathbb{E}[\xi_1 \xi_2]
= \rho$. 

The characterization of the empirical distribution of eigenvalues for
stochastic blockmodel graphs is of significant interest, but
there are only a few available results. In particular, \cite{Zhang}
and \cite{avrachenkov} derived the Stieltjes transform for the
limiting empirical distribution of the bulk eigenvalues for stochastic
blockmodel graphs, thereby showing that the empirical distribution of
the eigenvalues need not converge to a semicircle law. \cite{Zhang}
and \cite{avrachenkov} also considered the edge eigenvalues, but their
characterization depends upon inverting the Stieltjes transform and thus
currently does not yield the limiting distribution for these largest
eigenvalues. \cite{lei2014} derived the limiting distribution for the
largest eigenvalue of a centered and scaled version of
$\mathbf{A}$. More specifically, \cite{lei2014} showed that there is a
consistent estimate $\widehat{\mathbb{E}[\mathbf{A}]} =
(\hat{a}_{ij})$ of $\mathbb{E}[\mathbf{A}]$ such that the matrix
$\tilde{\mathbf{A}} = (\tilde{a}_{ij})$ with entries $\tilde{a}_{ij} =
(a_{ij} - \hat{a}_{ij})/\sqrt{(n-1)\hat{a}_{ij} (1 - \hat{a}_{ij}}$
has a limiting Tracy-Widom distribution, i.e.,
$n^{2/3}(\lambda_{1}(\tilde{\mathbf{A}}) - 2)$ converges to 
Tracy-Widom.

This paper addresses the open question of determining the limiting
distribution of the edge eigenvalues of adjacency matrices for
stochastic blockmodel graphs. In particular, we extend the result of
F\"{u}redi and Koml\'{o}s and show that, in the limit, these
eigenvalues are jointly multivariate normal with bounded covariances.

\section{Main results} We present our result in the more general
framework of generalized random dot product graph where
$\mathbb{E}[\mathbf{A}]$ is only assumed to be low rank, i.e, we do
not require that the entries of $\mathbb{E}[\mathbf{A}]$ takes on a
finite number of distinct values.  We first define the notion of a
(generalized) random dot product graph
\citep{young2007random,rubin_delanchy_grdpg}.
\begin{definition}[Generalized random dot product graph]
\label{def:grdpg}
  Let $d$ be a positive integer and $p \geq 1$ and $q \geq 0$ be non-negative
  integers such that $p + q = d$. Let $\mathbf{I}_{p,q}$ denote the
  diagonal matrix whose first $p$ diagonal elements equal $1$ and the remaining $q$ diagonal
  entries equal $-1$. 
  Let $\mathcal{X}$ be a subset of 
  $\mathbb{R}^{d}$ such that $x^{\top} \mathbf{I}_{p,q} y \in[0,1]$ for all $x,y\in
  \mathcal{X}$. Let $F$ be a distribution taking values in $\mathcal{X}$. 
  We say $(\mathbf{X},\mathbf{A}) \sim
  \mathrm{GRDPG}(F)$ with signature $(p,q)$ if the following holds. First let $X_1, X_2,
  \dots, X_n \overset{\mathrm{i.i.d}}{\sim} F$ and set
  $\mathbf{X}=[X_1 \mid \cdots \mid X_n]^\top\in \mathbb{R}^{n\times
    d}$. Then $\mathbf{A}\in\{0,1\}^{n\times n}$ is a symmetric matrix
such that the entries $\{a_{ij}\}_{i \leq j}$ are independent and
\begin{equation}
 a_{ij} \sim \mathrm{Bernoulli}(X_i^\top \mathbf{I}_{p,q} X_j).
\end{equation}
We therefore have
\begin{equation}
\Pr[\mathbf{A} \mid \mathbf{X}]=\prod_{i \leq j} (X^{\top}_i \mathbf{I}_{p,q}
X_j)^{a_{ij}}(1- X^{\top}_i \mathbf{I}_{p,q} X_j)^{(1-a_{ij})}.
\end{equation}
When $q = 0$, we say that $(\mathbf{A}, \mathbf{X}) \sim \mathrm{RDPG}(F)$, i.e., 
$\mathbf{A}$ is a random dot product graph.
\end{definition}

\begin{remark}
Any stochastic blockmodel graph
$(\tau, \mathbf{A}) \sim \mathrm{SBM}(\bm{\pi}, \mathbf{B})$ can 
be represented as a (generalized) random dot product graph
$(\mathbf{X}, \mathbf{A}) \sim \mathrm{GRDPG}(F)$ where $F$ is a
mixture of point masses. Indeed, suppose $\mathbf{B}$ is a $K \times
K$ matrix and let $\mathbf{B} = \mathbf{U} \bm{\Sigma}
\mathbf{U}^{\top}$ be the eigendecomposition of $\mathbf{B}$. Then,
denoting by $\nu_1, \nu_2, \dots, \nu_K$ the rows of $\mathbf{U}
|\bm{\Sigma}|^{1/2}$, we can define $F = \sum_{k=1}^{K} \pi_k
\delta_{\nu_k}$ where $\delta$ is the Dirac delta function. The
signature $(p,q)$ is given by the number of positive and negative
eigenvalues of $\mathbf{B}$, respectively. 
Similar constructions show that degree-corrected stochastic blockmodel graphs \citep{karrer2011stochastic} and mixed-membership stochastic blockmodel graphs \citep{Airoldi2008} are also special cases of generalized random dot product graphs.  
\end{remark}

\begin{remark} We note that non-identifiability is
an intrinsic property of generalized random dot product graphs. More
specifically, if $(\mathbf{X}, \mathbf{A}) \sim \mathrm{GRDPG}(F)$
where $F$ is a distribution on $\mathbb{R}^{d}$ with signature $(p,q)$, then for any
matrix $\mathbf{W}$ such that $\mathbf{W}
\mathbf{I}_{p,q} \mathbf{W}^{\top} = \mathbf{I}_{p,q}$, we have that
$(\mathbf{Y}, \mathbf{B}) \sim
\mathrm{RDPG}(F \circ \mathbf{W})$ is identically distributed to $(\mathbf{X},
\mathbf{A})$, where $F \circ \mathbf{W}$ denote the distribution
of $\mathbf{W} \xi$ 
for $\xi \sim F$. A matrix $\mathbf{W}$ satisfying $\mathbf{W}
\mathbf{I}_{p,q} \mathbf{W}^{\top} = \mathbf{I}_{p,q}$ is said to be
an {\em indefinite orthogonal} matrix. For the special case of random
dot product graphs where $q = 0$, the condition on $\mathbf{W}$ reduces to
that of an orthogonal matrix.
\end{remark}

With the above notations in place, we now state our generalization of
\cite{furedi1981eigenvalues} for the generalized random dot product graph setting.
\begin{theorem}
\label{THM:CLT_EIGENVAL} Let $(\mathbf{A}, \mathbf{X}) \sim
\mathrm{GRDPG}(F)$ be a $d$-dimensional generalized random dot product graph
with signature $(p,q)$. Let $\Delta = \mathbb{E}[X X^{\top}]$ where $X \sim F$
and suppose that $\Delta \mathbf{I}_{p,q}$ has $p+q = d$ simple eigenvalues. 
Let $\mathbf{P} = \mathbf{X}
\mathbf{I}_{p,q} \mathbf{X}^{\top}$ and for $1 \leq i \leq d$, let
$\hat{\lambda}_{i}$ and $\lambda_i$ be the $i$-th
largest eigenvalues of $\mathbf{A}$ and $\mathbf{P}$ (in modulus), respectively.
Let $\lambda_i(\Delta \mathbf{I}_{p,q})$ and $\bm{\xi}_i$ denote the $i$-th largest
eigenvalue and associated (unit-norm) eigenvector pair for the matrix 
$\Delta^{1/2} \mathbf{I}_{p,q} \Delta^{1/2}$. Let $\mu =\mathbb{E}[X]$ and denote by 
$\bm{\eta}$ the $d \times 1$ vector whose elements are
\begin{equation} 
\label{eq:def_mui}
\eta_i = \frac{1}{\lambda_i(\Delta \mathbf{I}_{p,q})} \mathbb{E}[\bm{\xi}_i^{\top}
  \Delta^{-1/2} X X^{\top} \Delta^{-1/2} \bm{\xi}_i (X^{\top}
\mathbf{I}_{p,q} \mu - X^{\top} \mathbf{I}_{p,q} \Delta
\mathbf{I}_{p,q} X)]
\end{equation}

Also let $\bm{\Gamma}$ be the $d \times d$ matrix whose elements are
\begin{equation}
\label{eq:def_gamma_ij}
\begin{split}
\Gamma_{ij} & = 2 \bigl(\mathbb{E}[\bm{\xi}_i^{\top} \Delta^{-1/2} X
X^{\top} \Delta^{-1/2} \bm{\xi}_j X]^{\top}
\mathbf{I}_{p,q} \mathbb{E}[\bm{\xi}_i^{\top} \Delta^{-1/2} X
X^{\top} \Delta^{-1/2} \bm{\xi}_j
X]\bigr)
\\
& - 
 2
\mathrm{tr} \bigl(\mathbb{E}[\bm{\xi}_i^{\top} \Delta^{-1/2} X
X^{\top} \Delta^{-1/2} \bm{\xi}_j X X^{\top}]
\mathbf{I}_{p,q} \mathbb{E}[\bm{\xi}_i^{\top} \Delta^{-1/2} X
X^{\top} \Delta^{-1/2} \bm{\xi}_j X
X^{\top}] \mathbf{I}_{p,q}\bigr)
\end{split}
 \end{equation}
We then have
$$ (\hat{\lambda}_1- \lambda_1, \hat{\lambda}_2 - \lambda_2, 
\dots, \hat{\lambda}_d - \lambda_{d}) \overset{\mathrm{d}}{\longrightarrow} \mathrm{MVN}(\bm{\eta}, \bm{\Gamma}) $$
as $n \rightarrow \infty$.
\end{theorem}
When $\mathbf{A}$ is a $d$-dimensional random dot product graph, 
Theorem~\ref{THM:CLT_EIGENVAL} simplifies to the following result. 
\begin{corollary}
  \label{cor:CLT_rdgp}
  Let $(\mathbf{A}, \mathbf{X}) \sim
\mathrm{RDPG}(F)$ be a $d$-dimensional random dot product graph
and suppose that $\Delta = \mathbb{E}[ X X^{\top}]$ has $d$ simple eigenvalues. 
Let $\mathbf{P} = \mathbf{X} \mathbf{X}^{\top}$ and let 
$\lambda_i(\Delta)$ and $\bm{\xi}_i$ denote the $i$-th largest
eigenvalue and associated (unit-norm) eigenvector of $\Delta$. 
Let $\mu =\mathbb{E}[X]$ and denote by $\bm{\eta}$ the $d \times 1$ vector with
elements 
\begin{equation} 
\label{eq:def_mui2}
\eta_i = \frac{\mathbb{E}[\bm{\xi}_i^{\top} X X^{\top} \bm{\xi}_i
  (X^{\top} \mu - X^{\top} \Delta X)]}
{\lambda_i(\Delta)^{2}}. 
\end{equation}
and by $\bm{\Gamma}$ the $d \times d$ matrix whose elements are
\begin{equation}
\label{eq:def_gamma_ij2}
\begin{split}
\Gamma_{ij} & = \frac{2}{\lambda_i(\Delta) \lambda_j(\Delta)}
\bigl(\mathbb{E}[\bm{\xi}_i^{\top} X X^{\top} \bm{\xi}_j X]^{\top}
\mathbb{E}[\bm{\xi}_i^{\top} X X^{\top} \bm{\xi}_j X]\bigr) \\ &-
\frac{2}{\lambda_i(\Delta) \lambda_j(\Delta)} \mathrm{tr}
\bigl(\mathbb{E}[\bm{\xi}_i^{\top} X X^{\top} \bm{\xi}_j X X^{\top}]
 \mathbb{E}[\bm{\xi}_i^{\top} X X^{\top} \bm{\xi}_j X X^{\top}]\bigr)
 \end{split}
 \end{equation}
We then have
$$ (\hat{\lambda}_1 - \lambda_1, \hat{\lambda}_2 - 
\lambda_2, \dots, \hat{\lambda}_{d}- \lambda_{d}) \longrightarrow \mathrm{MVN}(\bm{\eta}, \bm{\Gamma}) $$
as $n \rightarrow \infty$.
\end{corollary}
To illustrate Corollary~\ref{cor:CLT_rdgp}, let $\mathbf{A}$ be an
Erd\H{o}s-R\'{e}nyi graph with edge probability $p$; then $F$ is the
Dirac delta measure at $p^{1/2}$ and hence $\Delta = p$, $\bm{\xi}_1 =
1$, and $\lambda_i(\Delta) = p$. We thus recover the earlier result of F\"{u}redi and
Koml\'{o}s that
$$ \hat{\lambda}_i - np \longrightarrow \mathcal{N}(1-p, 2p(1-p)).$$

\begin{figure}[tp!]
  \centering
  \includegraphics[width=0.7\textwidth]{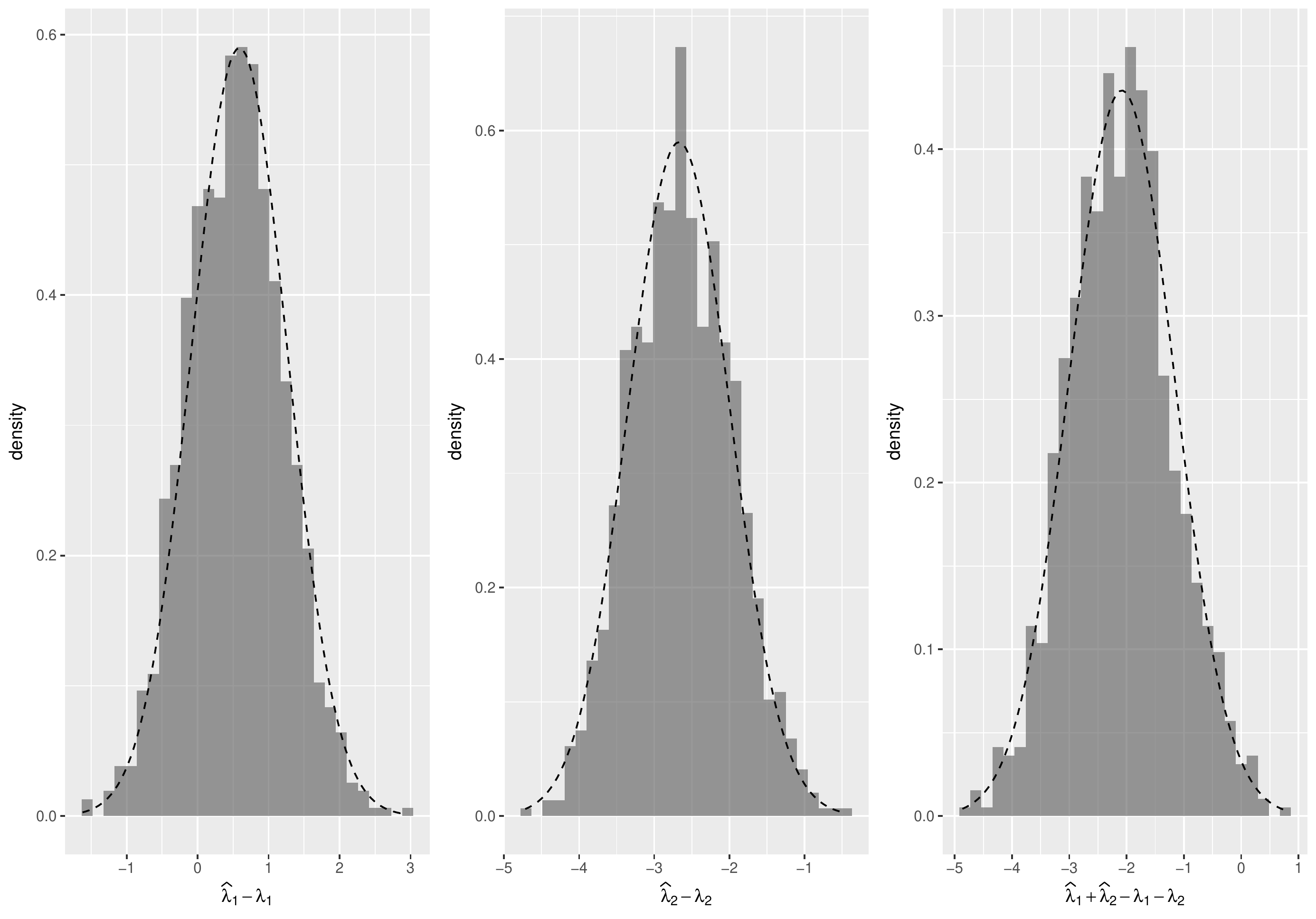}
  \caption{Plot of the empirical distribution for
    $\hat{\lambda}_1 - \lambda_1$,
    $\hat{\lambda}_2 - \lambda_2$, and $\hat{\lambda}_1 +
    \hat{\lambda}_2 - \lambda_1 - \lambda_2$ for $1000$ samples
    of two-block SBM graphs on $n = 4000$ vertices. The SBM parameters
    are $\mathbf{B} = \left(\protect\begin{smallmatrix} 0.3 & 0.5 \\ 0.5 &
      0.3 \protect\end{smallmatrix}\right)$
    and $\bm{\pi} = (0.3,0.7)$. The dashed lines are the
    probability density function for the normal distribution
    with parameters specified as in Theorem~\ref{THM:CLT_EIGENVAL}.
  }
\label{plot:grdpg_rk2_plot1}
\end{figure}

When the eigenvalues of $\Delta \mathbf{I}_{p,q}$ are not all 
simple eigenvalues, Theorem~\ref{THM:CLT_EIGENVAL} can be adapted to yield the
following result. 
\begin{theorem}
\label{THM:ARBITRARY} Let $(\mathbf{X}, \mathbf{A}) \sim
\mathrm{GRDPG}(F)$ be a $d$-dimensional generalized random dot
product graph on $n$ vertices with signature $(p,q)$. Let $\mathbf{P} = \mathbf{X} \mathbf{I}_{p,q}
\mathbf{X}^{\top}$ and for $1 \leq i \leq r$, let $\hat{\lambda}_{i}$
and $\lambda_i$ denote the $i$-th largest eigenvalues of $\mathbf{A}$
and $\mathbf{P}$ (in modulus), respectively. Also let $\bm{v}_i$ be the
unit norm eigenvector satisfying $(\mathbf{X}^{\top}
\mathbf{X})^{1/2} \mathbf{I}_{p,q} (\mathbf{X}^{\top}
\mathbf{X})^{1/2} \bm{v}_i = \lambda_i \bm{v}_i$ for $i
=1,2,\dots,d$. Denote by $\tilde{\bm{\eta}} = \tilde{\bm{\eta}}(\mathbf{X})$ the $d
\times 1$ vector with elements
\begin{equation} 
\label{eq:def_mui3}
\tilde{\eta}_i = 
\tfrac{n}{\lambda_i} \Bigl(\tfrac{1}{n} \sum_{s=1}^{n} \bm{v}_i^{\top} (\tfrac{\mathbf{X}^{\top}
\mathbf{X}}{n})^{-1/2} X_s X_s^{\top} (\tfrac{\mathbf{X}^{\top}
\mathbf{X}}{n})^{-1/2} \bm{v}_i X_s^{\top} \mathbf{I}_{p,q}
\bigl(\tfrac{1}{n} \sum_{t=1}^{n} (X_t - X_t X_t^{\top} \mathbf{I}_{p,q} X_s)
\bigr) \Bigr)
\end{equation}
and by $\bm{\sigma}^{2} = \bm{\sigma}^{2}(\mathbf{X})$ the $d \times 1$ vector whose elements are
\begin{equation}
\label{eq:def_gamma_ij3}
\begin{split}
 \sigma^{2}_{i} &= 2 \Bigl(\sum_{k} (X_k^{\top} (\mathbf{X}^{\top} \mathbf{X})^{-1/2}
\bm{v}_i)^2 X_k^{\top} \Bigr) \mathbf{I}_{p,q} \Bigl(\sum_{l}
(X_l^{\top} (\mathbf{X}^{\top} \mathbf{X})^{-1/2}
\bm{v}_i)^2 \Bigr) \\ & - 2 \mathrm{tr}
\Bigl(\sum_{k} (X_k^{\top} (\mathbf{X}^{\top} \mathbf{X})^{-1/2}
\bm{v}_i)^2   X_k X_k^{\top} \Bigr) \mathbf{I}_{p,q} \Bigl(\sum_{l}
(X_l^{\top} (\mathbf{X}^{\top} \mathbf{X})^{-1/2} \bm{v}_i)^2 
X_l X_l^{\top} \Bigr) \mathbf{I}_{p,q} 
 \end{split}
 \end{equation}
We then have, for $1 \leq i \leq d$, 
$$ \frac{1}{\sigma_{i}} (\hat{\lambda}_i - \lambda_i - \tilde{\eta}_i)
\longrightarrow \mathrm{N}(0, 1) $$
as $n \rightarrow \infty$.
\end{theorem}
The main differences between Theorem~\ref{THM:ARBITRARY} and 
Theorem~\ref{THM:CLT_EIGENVAL} are that (1) we do not claim that the quantities
$\tilde{\eta}_i$ and $\sigma_{i}^2$ in Theorem~\ref{THM:ARBITRARY} (which, for $(\mathbf{X},
\mathbf{A}) \sim \mathrm{GRDPG}(F)$ are functions of the
underlying latent positions $\mathbf{X}$) converge as $n \rightarrow \infty$ and (2) we do not claim that the
collection $(\hat{\lambda}_i - \lambda_i)_{i=1}^{d}$ in
Theorem~\ref{THM:ARBITRARY} converges jointly to multivariate
normal. The above diffences stem mainly from the fact that when the
eigenvalues of $\Delta \mathbf{I}_{p,q}$ are not simple eigenvalues, 
then $\tfrac{\mathbf{X}^{\top} \mathbf{X}}{n} \rightarrow \Delta$ as $n \rightarrow \infty$ but $\bm{v}_i$ does not necessarily converges to $\bm{\xi}_i$, the corresponding eigenvector of $\Delta^{1/2}
\mathbf{I}_{p,q} \Delta^{1/2}$, as $n \rightarrow \infty$.
 
\begin{figure}[tp!]
  \centering
  \includegraphics[width=0.6\textwidth]{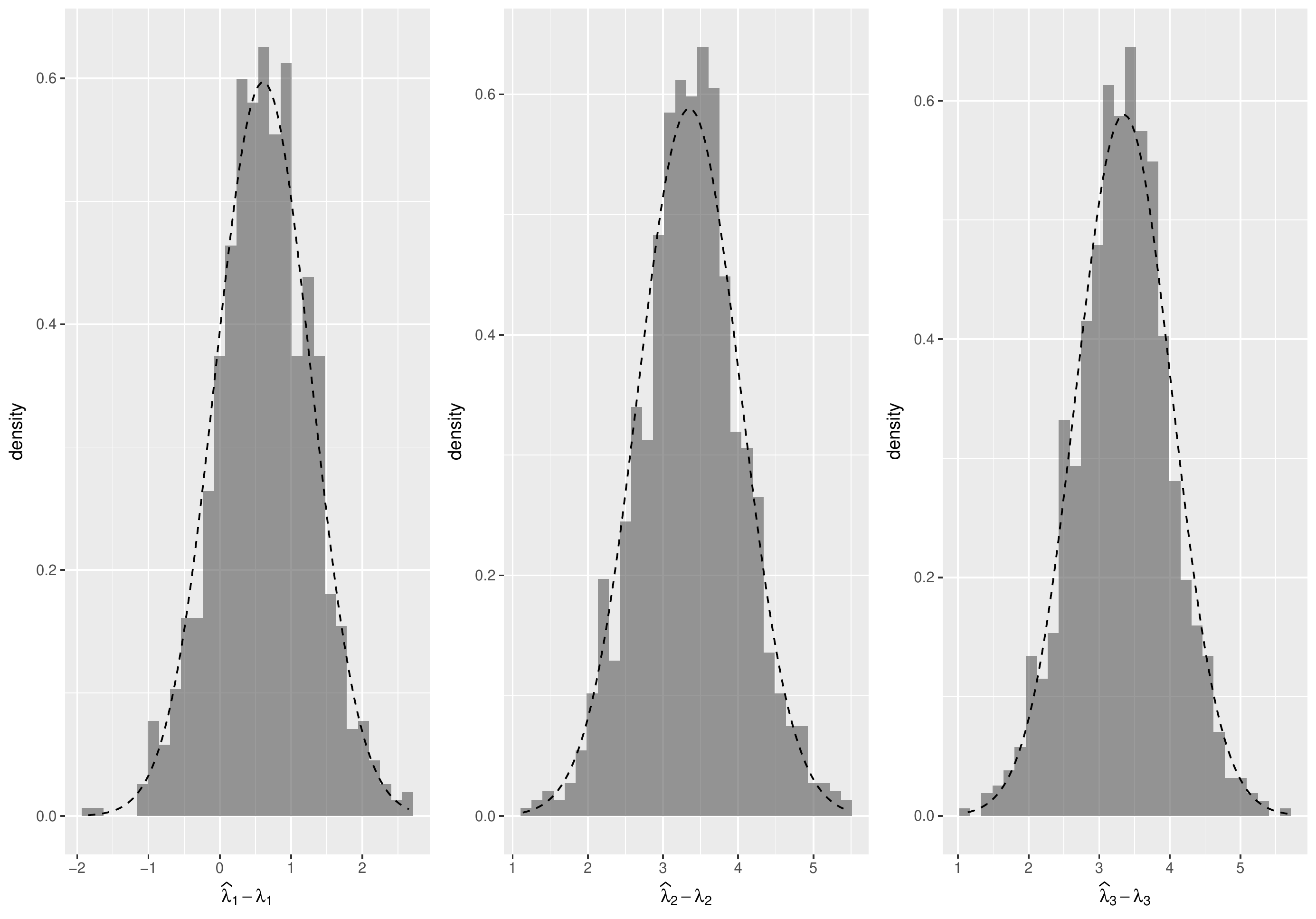}
  \caption{Plot of the empirical distribution for
    $\hat{\lambda}_1 - \lambda_1$,
    $\hat{\lambda}_2 - \lambda_2$, and $\hat{\lambda}_3 - \lambda_3$ for $1000$ samples
    of a three-block SBM graphs on $n = 4000$ vertices. The SBM parameters
    are $\mathbf{B} = 0.2 \mathbf{I} + 0.3 \bm{1} \bm{1}^{\top}$
    and $\bm{\pi} = (\tfrac{1}{3},\tfrac{1}{3},\tfrac{1}{3})$. The
    dashed lines are the
    probability density function for the normal distributions
    with mean $\tilde{\eta}_i$ and variance $\sigma_{i}^2$ specified as in
    Theorem~\ref{THM:ARBITRARY}. Note that $\Delta$ has repeated
    eigenvalues, i.e., the eigenvalues of $\Delta$ are $11/30$, $2/30$
    and $2/30$.
  }
\label{plot:grdpg_rk3_plot1}
\end{figure}

\section{Proof of Theorem~\ref{THM:CLT_EIGENVAL} and Theorem~\ref{THM:ARBITRARY}}
Let $\bm{u}_1, \bm{u}_2,
\dots, \bm{u}_d$ be the eigenvectors corresponding to the non-zero
eigenvalues $\lambda_1, \lambda_2, \dots, \lambda_d$ of
$\mathbf{P}$. Similarly, let $\hat{\bm{u}}_1, \hat{\bm{u}}_2, \dots
\hat{\bm{u}}_d$ be the eigenvectors corresponding to the
eigenvalues $\hat{\lambda}_1, \hat{\lambda}_2, \dots, \hat{\lambda}_d$
of $\mathbf{A}$. 

A sketch of the proof of Theorem~\ref{THM:CLT_EIGENVAL} and Theorem~\ref{THM:ARBITRARY} is as
follows. First we derive the following approximation
of $\hat{\lambda}_i - \lambda_i$ by a
sum of two quadratic forms $\bm{u}_i^{\top}(\mathbf{A} - \mathbf{P})
\bm{u}_i$ and $\bm{u}_i^{\top} (\mathbf{A}
- \mathbf{P})^2 \bm{u}_i$, namely
\begin{equation}
\label{eq:main_decomp_proof}
\begin{split}
\hat{\lambda}_i - \lambda_i &=  
\frac{\lambda_i}{\hat{\lambda}_i} \bm{u}_i^{\top} (\mathbf{A} -
\mathbf{P}) \bm{u}_i + \frac{\lambda_i}{\hat{\lambda}_i^2}
\bm{u}_i^{\top} (\mathbf{A} - \mathbf{P})^2 \bm{u}_i + 
O_{\mathbb{P}}(n^{-1/2}).
\end{split} 
\end{equation}
Now, the term $\lambda_i^{-1} \bm{u}_i^{\top}(\mathbf{A} - \mathbf{P})^2
\bm{u}_i$ is a function of the $n(n+1)/2$ independent random variables
$\{a_{rs} - p_{rs}\}_{r \leq s}$ and hence is concentrated around its
expectation, i.e.,
\begin{equation}
  \label{eq:lln1}
\frac{\lambda_i}{\hat{\lambda}_i^2} \bm{u}_i^{\top}(\mathbf{A} - \mathbf{P})^2
\bm{u}_i = \mathbb{E}[\lambda_i^{-1} \bm{u}_i^{\top}(\mathbf{A} - \mathbf{P})^2
\bm{u}_i] + O_{\mathbb{P}}(n^{-1/2})
\end{equation}
where the expectation is taken with respect to
$\mathbf{A}$. Letting $\tilde{\eta}_i = \mathbb{E}[\lambda_i^{-1} \bm{u}_i^{\top}(\mathbf{A} - \mathbf{P})^2
\bm{u}_i]$, we obtain, after some straightforward algebraic manipulations, 
the expression for $\tilde{\eta}_i$ in Eq.~\eqref{eq:def_mui3}. 
When the eigenvalues of $\Delta \mathbf{I}_{p,q}$ are distinct, 
we derive the limit $\tilde{\eta}_i \overset{\mathrm{a.s.}}{\longrightarrow} \eta_i$ where $\eta_i$ is defined in Eq.~\eqref{eq:def_mui}. 
Next, with $u_{is}$
being the $s$-th element of $\bm{u}_i$,
$$\bm{u}_i^{\top}(\mathbf{A} - \mathbf{P})
\bm{u}_i = \sum_{r < s} 2 u_{is} u_{ir} (a_{rs} - p_{rs}) + \sum_{r}
u_{ir}^2 (a_{rr} - p_{rr})$$ 
is, conditional on $\mathbf{X}$, a sum of independent mean $0$ random
variables and the Lindeberg-Feller central limit theorem yield
\begin{equation}
  \label{eq:u.t(A-P)u_normal}
 \frac{\lambda_i}{\hat{\lambda}_i \sigma_{i}} \bm{u}_i^{\top}(\mathbf{A} - \mathbf{P})
\bm{u}_i \overset{\mathrm{d}}{\longrightarrow} \mathcal{N}(0, 1)
\end{equation}
as $n \rightarrow \infty$, where $\sigma_{i}^2$ is as defined in Eq.~\eqref{eq:def_gamma_ij3}. 
Thus for each $i \leq d$, 
$\tfrac{1}{\sigma_i} (\lambda_i - \hat{\lambda}_i - \tilde{\eta}_i) \rightarrow N(0, 1)$ as $n
\rightarrow \infty$. When the eigenvalues of $\Delta \mathbf{I}_{p,q}$ are distinct, then $\sigma^{2}_{i} \overset{\mathrm{a.s.}}{\longrightarrow} \Gamma_{ii}$ as defined in Eq.~\eqref{eq:def_gamma_ij}. The joint distribution of $(\hat{\lambda}_i - \lambda_i)_{i=1}^{d}$ in Theorem~\ref{THM:CLT_EIGENVAL} then follows from the Cramer-Wold device. 

We now provide detailed derivations 
of Eq.~\eqref{eq:main_decomp_proof} through
Eq.~\eqref{eq:u.t(A-P)u_normal}.

{\bf Proof of Eq.~\eqref{eq:main_decomp_proof}}
For a given $i \leq d$, we have
\begin{equation*}
\begin{split}
\bigl(\hat{\lambda}_i \mathbf{I} - (\mathbf{A} - \mathbf{P})\bigr) \hat{\bm{u}}_i = \mathbf{A} \hat{\bm{u}}_i - (\mathbf{A} - \mathbf{P}) \hat{\bm{u}}_i = \mathbf{P} \hat{\bm{u}}_i = \bigl(\sum_{j=1}^{r} \lambda_j \bm{u}_j \bm{u}_j^{\top}\bigr)  \hat{\bm{u}}_i
\end{split}
 \end{equation*}
 Now suppose that $\hat{\lambda}_i \mathbf{I} - (\mathbf{A} -
 \mathbf{P})$ is invertible; this holds with high probability
 for sufficiently large $n$. Then multiplying both sides of the above display by $\bm{u}_i^{\top}
 (\hat{\lambda}_i \mathbf{I} - (\mathbf{A} - \mathbf{P}))^{-1}$ on the
 left and using the von Neumann identity $(\mathbf{I} -
 \mathbf{X})^{-1} = \mathbf{I} + \sum_{k=1}^{\infty} \mathbf{X}^{k}$ for $\|\mathbf{X}\| < 1$, we have
\begin{equation}
\label{eq:vonNeumann}
\begin{split}
\bm{u}_i^{\top} \hat{\bm{u}}_i &= \sum_{j=1}^{d} \lambda_j \bm{u}_i^{\top} (\hat{\lambda}_i \mathbf{I} - (\mathbf{A} - \mathbf{P}))^{-1} \bm{u}_j \bm{u}_j^{\top} \hat{\bm{u}}_i = \sum_{j=1}^{d} \lambda_j \hat{\lambda}_i^{-1} \bm{u}_i^{\top} (\mathbf{I} - \hat{\lambda}_i^{-1}(\mathbf{A} - \mathbf{P}))^{-1} \bm{u}_j \bm{u}_j^{\top} \hat{\bm{u}}_i
\\ &= \sum_{j=1}^{d} \lambda_j \hat{\lambda}_i^{-1} \bm{u}_i^{\top} \Bigl(\mathbf{I} + \sum_{k=1}^{\infty} \hat{\lambda}_i^{-k} (\mathbf{A} - \mathbf{P})^{k} \Bigr) \bm{u}_j \bm{u}_j^{\top} \hat{\bm{u}}_i 
\\ &=
\frac{\lambda_i}{\hat{\lambda}_i} \bm{u}_i^{\top} \Bigl(\mathbf{I} + \sum_{k=1}^{\infty} \hat{\lambda}_i^{-k} (\mathbf{A} - \mathbf{P})^{k} \Bigr) \bm{u}_i \bm{u}_i^{\top} \hat{\bm{u}}_i + \sum_{j \not = i} \frac{\lambda_j}{\hat{\lambda}_i} \bm{u}_i^{\top} \Bigl(\sum_{k=1}^{\infty} \hat{\lambda}_i^{-k} (\mathbf{A} - \mathbf{P})^{k} \Bigr) \bm{u}_j \bm{u}_j^{\top} \hat{\bm{u}}_i
\end{split}
\end{equation} 
We first assume that all of the eigenvalues of $\Delta \mathbf{I}_{p,q}$ are simple eigenvalues. The eigenvalues of $\mathbf{P} = \mathbf{X}^{\top} \mathbf{I}_{p,q} \mathbf{X}^{\top}$ are then well-separated, i.e., 
$\min_{j \not = i} |\lambda_i - \lambda_j| = O_{\mathbb{P}}(n)$ for $1 \leq i \not = j \leq d$.
The Davis-Kahan $\sin \Theta$ theorem \citep{davis70,samworth} therefore
implies, for some constant $C$,
\begin{gather}
\label{eq:u_i1}
1 - \bm{u}_i^{\top} \hat{\bm{u}}_i = 
\tfrac{1}{2} \|\bm{u}_i - \hat{\bm{u}}_i \|^2 \leq \frac{C^2 \|\mathbf{A} - \mathbf{P}\|^2}{\min\{|\lambda_i - \lambda_{i+1}|^{2},|\lambda_{i-1} - \lambda_i|^{2}\}} =
O_{\mathbb{P}}(n^{-1}) \\
\label{eq:u_i2}
|\bm{u}_j^{\top} \hat{\bm{u}}_i| \leq \frac{C \|\mathbf{A} - \mathbf{P}\|}{\min\{|\lambda_i - \lambda_{i+1}|,|\lambda_{i-1} - \lambda_i|\} 
} = O_{\mathbb{P}}(n^{-1/2}).
\end{gather}
We can thus divide both side of the above display by $\bm{u}_i^{\top}
\hat{\bm{u}}_i$ to obtain
\begin{equation*}
\begin{split}
1 = \frac{\lambda_i}{\hat{\lambda}}_i +
\frac{\lambda_i}{\hat{\lambda}_i} \bm{u}_i^{\top} 
\Bigl(\sum_{k=1}^{\infty} \hat{\lambda}_i^{-k} (\mathbf{A} -
\mathbf{P})^{k} \Bigr) 
\bm{u}_i +
\sum_{j \not = i} \frac{\lambda_j}{\hat{\lambda}_i}
\bm{u}_i^{\top} 
\Bigl(\sum_{k=1}^{\infty} \hat{\lambda}_i^{-k} (\mathbf{A} -
\mathbf{P})^{k} \Bigr)
\bm{u}_j \frac{\bm{u}_j^{\top} \hat{\bm{u}}_i}{\bm{u}_i^{\top} \hat{\bm{u}}_i}.
\end{split}
\end{equation*}
Equivalently, 
\begin{equation}
\label{eq:divide_u_i_top}
\begin{split}
\hat{\lambda}_i - \lambda_i =  \lambda_i \bm{u}_i^{\top}
\Bigl(\sum_{k=1}^{\infty} \hat{\lambda}_i^{-k} (\mathbf{A} -
\mathbf{P})^{k} \Bigr) \bm{u}_i  
&+ \sum_{j \not = i} 
\frac{\lambda_j}{\hat{\lambda}_i} \bm{u}_i^{\top} (\mathbf{A} -
\mathbf{P})
 \bm{u}_j \frac{\bm{u}_j^{\top} \hat{\bm{u}}_i}{\bm{u}_i^{\top} \hat{\bm{u}}_i} \\
&+ \sum_{j \not = i} \lambda_j \bm{u}_i^{\top}
\Bigl(\sum_{k=2}^{\infty} 
\hat{\lambda}_i^{-k} (\mathbf{A} - \mathbf{P})^{k} \Bigr) \bm{u}_j \frac{\bm{u}_j^{\top} \hat{\bm{u}}_i}{\bm{u}_i^{\top} \hat{\bm{u}}_i}.
\end{split}
\end{equation} 
Now $\lambda_i^{-1} \hat{\lambda}_j = O_{\mathbb{P}}(1)$, and by
Hoeffding's inequality, $\bm{u}_j^{\top} (\mathbf{A} -
\mathbf{P}) \bm{u}_i = O_{\mathbb{P}}(1)$. Since $\bm{u}_j^{\top}
\hat{\bm{u}}_i = O_{\mathbb{P}}(n^{-1/2})$, we have
$$ \sum_{j \not = i} 
\frac{\lambda_j}{\hat{\lambda}_i} \bm{u}_i^{\top} 
(\mathbf{A} - \mathbf{P}) \bm{u}_j \frac{\bm{u}_j^{\top} \hat{\bm{u}}_i}{\bm{u}_i^{\top} \hat{\bm{u}}_i} = O_{\mathbb{P}}(n^{-1/2}).$$
Next we note that $\|\sum_{k=2}^{\infty} \hat{\lambda}_i^{-k} (\mathbf{A} - \mathbf{P})^{k} \|$ can be bounded as
\begin{equation}
  \label{eq:power_series1}
  \|\sum_{k=2}^{\infty} \hat{\lambda}_i^{-k} (\mathbf{A} -
\mathbf{P})^{k} \| \leq \frac{\|\hat{\lambda}_i^{-2} 
(\mathbf{A} - \mathbf{P})^2 \|}{1 - \|\hat{\lambda}_i^{-1} (\mathbf{A} - \mathbf{P}) \|} = O_{\mathbb{P}}(\hat{\lambda}_i^{-1}).
\end{equation}
We thus have
$$
\sum_{j \not = i} \lambda_j \bm{u}_i^{\top} \Bigl(\sum_{k=2}^{\infty}
\hat{\lambda}_i^{-k} (\mathbf{A} - \mathbf{P})^{k} \Bigr) \bm{u}_j 
\frac{\bm{u}_j^{\top} \hat{\bm{u}}_i}{\bm{u}_i^{\top} \hat{\bm{u}}_i} = O_{\mathbb{P}}(n^{-1/2})
$$
The above bounds then implies
\begin{equation}
\label{eq:diff_lambda}
\begin{split}
\hat{\lambda}_i - \lambda_i &=  \lambda_i \bm{u}_i^{\top} \Bigl(\sum_{k=1}^{\infty} \hat{\lambda}_i^{-k} (\mathbf{A} - \mathbf{P})^{k} \Bigr) \bm{u}_i + O_{\mathbb{P}}(n^{-1/2}).
\end{split}
\end{equation}
Similar to the derivation of Eq.~\eqref{eq:power_series1}, we also
show that
\begin{equation} 
\label{eq:power-A-P}
\|\sum_{k=3}^{\infty} \hat{\lambda}_i^{-k} (\mathbf{A} - \mathbf{P})^{k} \| \leq C \|\hat{\lambda}_i^{-3} (\mathbf{A} - \mathbf{P})^3 \| \leq C \hat{\lambda}_i^{-3/2} 
\end{equation}
with high probability
and thus Eq.~\eqref{eq:diff_lambda} and Eq.~\eqref{eq:power-A-P} imply
\begin{equation}
\label{eq:diff1}
\begin{split}
\hat{\lambda}_i - \lambda_i &=  
\frac{\lambda_i}{\hat{\lambda}_i} \bm{u}_i^{\top} (\mathbf{A} -
\mathbf{P}) \bm{u}_i + \frac{\lambda_i}{\hat{\lambda}_i^2}
\bm{u}_i^{\top} (\mathbf{A} - \mathbf{P})^2 \bm{u}_i + 
O_{\mathbb{P}}(n^{-1/2}),
\end{split} 
\end{equation}
and Eq.~\eqref{eq:main_decomp_proof} is established.

We now consider the case where the $i$-th eigenvalue of $\Delta \mathbf{I}_{p,q}$ has multiplicity $r_i \geq 2$. Let $\mathcal{S}_i$ be the indices of the $r_i$ eigenvalues $\lambda_j$ of $\mathbf{P} = \mathbf{X} \mathbf{I}_{p,q} \mathbf{X}^{\top}$ that is closest to $n \lambda_i(\Delta \mathbf{I}_{p,q})$, i.e.,
$$ \max_{j \in \mathcal{S}_i} |\lambda_j - n \lambda_i(\Delta \mathbf{I}_{p,q})| \leq \min_{k \not \in \mathcal{S}_i} |\lambda_k - n \lambda_i(\Delta \mathbf{I}_{p,q})|; \quad |\mathcal{S}_i| = r_i.$$
Denote by $\mathbf{U}_{\mathcal{S}_i}$ the $n \times r_i$ matrix whose columns are the eigenvectors corresponding to the $\lambda_j, j \in \mathcal{S}_i$. We note that $i \in \mathcal{S}_i$ with high probability for sufficiently large $n$. Furthermore, $|\lambda_j - \lambda_k| = O_{\mathbb{P}}(n)$ for all $j \in \mathcal{S}_i$ and $k \not \in \mathcal{S}_i$. Therefore, by the Davis-Kahan theorem, $\bm{u}_i^{\top} \hat{\bm{u}}_k = O_{\mathbb{P}}(n^{-1/2})$ for all $k \not \in \mathcal{S}_i$. We now consider $\bm{u}_i^{\top} \hat{\bm{u}}_j$ for $j \in \mathcal{S}_i, j \not = i$. We note that
\begin{equation} 
\begin{split}
\bm{u}_i^{\top} \hat{\bm{u}}_j &= \frac{\bm{u}_i^{\top} \mathbf{A} \hat{\bm{u}}_j - \bm{u}_i^{\top} \mathbf{P} \hat{\bm{u}}_j}{\hat{\lambda}_j - \lambda_i} = \frac{\bm{u}_i^{\top} (\mathbf{A} - \mathbf{P}) \mathbf{U}_{\mathcal{S}_i} \mathbf{U}_{\mathcal{S}_i}^{\top} \hat{\bm{u}}_j}{\hat{\lambda}_j - \lambda_i} + \frac{\bm{u}_i^{\top} (\mathbf{A} - \mathbf{P}) (\mathbf{I} - \mathbf{U}_{\mathcal{S}_i} \mathbf{U}_{\mathcal{S}_i}^{\top}) \hat{\bm{u}}_j}{\hat{\lambda}_j - \lambda_i} \\
&= \frac{n^{-1/2} \bm{u}_i^{\top} (\mathbf{A} - \mathbf{P}) \mathbf{U}_{\mathcal{S}_i} \mathbf{U}_{\mathcal{S}_i}^{\top} \hat{\bm{u}}_j}{n^{-1/2}(\hat{\lambda}_j - \lambda_j) + n^{-1/2}(\lambda_j - \lambda_i)} + \frac{n^{-1/2}\bm{u}_i^{\top} (\mathbf{A} - \mathbf{P}) (\mathbf{I} - \mathbf{U}_{\mathcal{S}_i} \mathbf{U}_{\mathcal{S}_i}^{\top}) \hat{\bm{u}}_j}{n^{-1/2}(\hat{\lambda}_j - \lambda_j) + n^{-1/2}(\lambda_j - \lambda_i)}.
\end{split}
\end{equation}
By Hoeffding inequality, $\bm{u}_i^{\top} (\mathbf{A} - \mathbf{P}) \mathbf{U}_{\mathcal{S}_i} = O_{\mathbb{P}}(1)$ with high probability. Since $j \in \mathcal{S}_i$, we have $\|(\mathbf{I} - \mathbf{U}_{\mathcal{S}_i} \mathbf{U}_{\mathcal{S}_i}^{\top}) \hat{\bm{u}}_j\| = O_{\mathbb{P}}(n^{-1/2})$ by the Davis-Kahan theorem. We then bound $\hat{\lambda}_j - \lambda_j$ using the following result of \citep[Theorem~3.7]{cape_16_conc} (see also \citet[Theorem~23]{orourke13:_random}).
\begin{theorem}
Let $\mathbf{A}$ and be a $n \times n$ symmetric random matrix with $\mathbf{A}_{ij} \sim \mathrm{Bernoulli}(\mathbf{P}_{ij})$ for $i \leq j$ and the entries $\{\mathbf{A}_{ij}\}$ are independent. Denote the $d+1$ largest singular values of $\mathbf{A}$ by $0 \leq \hat{\sigma}_{d+1} < \hat{\sigma}_d \leq \hat{\sigma}_{d-1} \leq \dots \leq \hat{\sigma}_1 $, and denote the $d+1$ largest singular values of $\mathbf{P}$ by $0 \leq \sigma_{d+1} < \sigma_d \leq \sigma_{d-1} \leq \dots \leq \sigma_1$. Suppose that $\Upsilon = \max_{i} \sum_{j} \mathbf{P}_{ij} = \omega(\log^{4}{n})$, $\sigma_{1} \geq C \Upsilon$, $\sigma_{d+1} \leq c \Upsilon$ for some absolute constants $C > c > 0$. Then for each $k \in \{1,2,\dots,d\}$, there exists some positive constant $c_{k,d}$ such that as $n \rightarrow \infty$, with probability at least $1 - n^{-3}$, we have
$$ |\hat{\sigma}_k - \sigma_k| \leq c_{k,d} \log{n}.$$
\end{theorem}
We thus have
$$|\bm{u}_i^{\top} \hat{\bm{u}}_j| = \frac{n^{-1/2} O_{\mathbb{P}}(1)}{n^{-1/2} O_{\mathbb{P}}(\log{n}) + n^{-1/2} (\lambda_j - \lambda_i)}.$$
We now analyze $n^{-1/2}(\lambda_j - \lambda_i)$. We can view $\mathbf{P} = \mathbf{X} \mathbf{I}_{p,q} \mathbf{X}^{\top}$ as a kernel matrix with symmetric kernel $h(X_i, X_j) = X_i^{\top} \mathbf{I}_{p,q} X_j$ where $X_i, X_j \sim F$. As $h$ is finite-rank, let $(\phi_i, \lambda_i(\mathbf{I}_{p,q} \Delta))_{i=1}^{d}$ denote the eigenvalues and associated eigenfunctions of the integral operator $\mathcal{K}_{h} \colon L_{2}(F) \mapsto L_{2}(F)$, i.e., 
$$\mathcal{K}_{h} \phi_i(x) \coloneqq \int h(x, y) \phi_i(y) \, \mathrm{d} F(y) = \lambda_i(\mathbf{I}_{p,q} \Delta) \phi_i(x).$$
Then, following \cite{koltchinskii00:_random}, let $\Psi_{i}$ denote the $r_i \times r_i$ random symmetric matrix whose half-vectorization $\mathrm{vech}(\Psi_i)$ is (jointly) distributed multivariate normal with mean $\bm{0}$ and $r_i(r_i+1)/2 \times r_i(r_{i}+1)/2$ covariance matrix with entries of the form 
$$\mathrm{Cov}(\Psi_{i}(s,t), \Psi_{i}(u,v)) = \int{\phi_s(y) \phi_t(y) \phi_{u}(y) \phi_v(y) \mathrm{d} F(y)}  - \int{\phi_s(y) \phi_t(y) \mathrm{d}F(y)} \int{\phi_u(y) \phi_v(y) \mathrm{d}F(y)}$$
for $1 \leq s \leq t \leq r_i, 1 \leq u \leq v \leq r_i$, where, with a slight abuse of notation, the collection $\{\phi_s\}_{s \leq r_i}$ denote the $r_i$ eigenfunctions of $\mathcal{K}_{h}$ associated with the eigenvalue $\lambda_i(\mathbf{I}_{p,q} \Delta)$.  
A simplification of the statement of Theorem~5.1 in \cite{koltchinskii00:_random}, to the setting where $h$ is a finite-rank kernel, yields
$$n^{1/2}(\lambda_j/n - \lambda_i(\mathbf{I}_{p,q} \Delta))_{j \in \mathcal{S}_i} \rightarrow \lambda_{i}(\mathbf{I}_{p,q} \Delta) \times (\lambda_s(\Psi_i))_{1 \leq s \leq r_i}$$
as $n \rightarrow \infty$; here we use the notation $\lambda_s(\mathbf{M})$ to denote the $s$-th largest eigenvalue, in modulus, of the matrix $\mathbf{M}$. 
Thus, the joint distribution of $\{n^{-1/2}(\lambda_s - n \lambda_{i}(\Delta \mathbf{I}_{p,q})\}_{s \in \mathcal{S}_i}$ converges to a non-degenerate limiting distribution and hence the limiting distribution of $n^{-1/2}(\lambda_i - \lambda_j)$ is also non-degenerate. 
We therefore have
$$ |\bm{u}_i^{\top} \hat{\bm{u}}_j| = \frac{n^{-1/2} O_{\mathbb{P}}(1)}{n^{-1/2} O_{\mathbb{P}}(\log{n}) + n^{-1/2} (\lambda_j - \lambda_i)} = o_{\mathbb{P}}(1); \quad j \in \mathcal{S}_i, j \not = i.$$ 
Finally, we note that there exists an orthogonal matrix $\mathbf{W}$ such that $\|\mathbf{U}^{\top} \hat{\mathbf{U}} - \mathbf{W}\| = O_{\mathbb{P}}(n^{-1})$. Hence, for any $i \leq d$,
$ \sum_{j=1}^{d} (\bm{u}_i^{\top} \hat{\bm{u}}_j)^2 = 1 + O_{\mathbb{P}}(n^{-1})$; hence, from our bounds for $\bm{u}_i^{\top} \hat{\bm{u}}_j$ for $j \not = i$ given above, we have $\bm{u}_i^{\top} \hat{\bm{u}}_i = 1 - o_{\mathbb{P}}(1)$. 
In summary, when the eigenvalues of $\mathbf{I}_{p,q} \Delta$ are not all simple eigenvalues, we have (in place of Eq.~\eqref{eq:u_i1} and Eq.~\eqref{eq:u_i2}), the bounds
\begin{equation}
\label{eq:u_i1_nodistinct}
 \bm{u}_i^{\top} \hat{\bm{u}}_i = 1 - o_{\mathbb{P}}(1); \quad \bm{u}_i^{\top} \hat{\bm{u}}_j = o_{\mathbb{P}}(1).
 \end{equation}
Thus Eq.\eqref{eq:divide_u_i_top} still holds and the remaining steps in the derivation of Eq.~\eqref{eq:diff1} can be easily adapted to yield
$$
\hat{\lambda}_i - \lambda_i =  
\frac{\lambda_i}{\hat{\lambda}_i} \bm{u}_i^{\top} (\mathbf{A} -
\mathbf{P}) \bm{u}_i + \frac{\lambda_i}{\hat{\lambda}_i^2}
\bm{u}_i^{\top} (\mathbf{A} - \mathbf{P})^2 \bm{u}_i + o_{\mathbb{P}}(1).$$


{\bf Proof of Eq.~\eqref{eq:lln1}} Let $Z = \lambda_i^{-1}
\bm{u}_i^{\top}(\mathbf{A} - \mathbf{P})^2 \bm{u}_i$. To derive
Eq.~\eqref{eq:lln1}, we show the concentration $Z$ around
$\mathbb{E}[Z]$ (where the expectation is taken with respect to
$\mathbf{A}$, conditional on $\mathbf{P}$) using a log-Sobolev
concentration inequality from \citet{boucheron2003}.  More
specifically, let $\mathbf{A}' = (a'_{rs})$ be an independent copy of
$\mathbf{A}$, i.e., the upper triangular entries of $\mathbf{A}'$ are
independent Bernoulli random variables with mean parameters $\{p_{rs}
\}_{r \leq s}$. For any pair of indices $(r,s)$, let
$\mathbf{A}^{(rs)}$ be the matrix obtained by replacing the $(r,s)$
and $(s,r)$ entries of $\mathbf{A}$ by $a'_{ij}$ and let $Z^{(rs)} =
\lambda_i^{-1} \bm{u}_i^{\top}(\mathbf{A}^{(rs)} - \mathbf{P})^2
\bm{u}_i$. Then Theorem~5 of \citet{boucheron13:_concen_inequal}
states that
\begin{theorem}
\label{THM:log_Sobolev}
Assume that there exists positive constants $a$ and $b$ such that 
$$ \sum_{r \leq s}(Z - Z^{(rs)})^2 \leq aZ + b. $$
Then for all $t > 0$,
\begin{gather} \mathbb{P}[Z - \mathbb{E}[Z] \geq t] \leq \exp\Bigl(\frac{-t^2}{4 a \mathbb{E}[Z] + 4b + 2at}\Bigr), \\
\mathbb{P}[Z - \mathbb{E}[Z] \leq -t] \leq \exp\Bigl(\frac{-t^2}{4 a \mathbb{E}[Z]} \Bigr).
\end{gather}
\end{theorem} The main technical step is then to bound $\sum_{r \leq
s}(Z - Z^{(rs)})^2$.  An identical argument to that in proving
Lemma~A.6 in \cite{tang14:_semipar} yield that
$$ \sum_{r \leq s}(Z - Z^{(rs)})^2 \leq a \lambda_i^{-1} Z + b. $$ 
for some constants $a$ and $b$. Theorem~\ref{THM:log_Sobolev} therefore implies 
\begin{equation*} 
\begin{split}
|Z - \mathbb{E}[Z]| \leq \sqrt{\mathbb{E}[Z]} \times O_{\mathbb{P}} (n^{-1/2})  = O_{\mathbb{P}}(n^{-1/2}).
\end{split}
\end{equation*}
as desired. 
 
We now evaluate $\mathbb{E}[Z] = \mathbb{E}[\lambda_i^{-1} \bm{u}_i^{\top}(\mathbf{A} - \mathbf{P})^2
\bm{u}_i]$. Let $\zeta_{rs}$ denote the $rs$-th entry of
$\mathbb{E}[(\mathbf{A} - \mathbf{P})^2]$. We note that $\zeta_{rs}$
is of the form
$$ \zeta_{rs} = \sum_{t} \mathbb{E}[(a_{rt} - p_{rt})(a_{st} - p_{st})] = 
\begin{cases} 0 & \text{if $r \not = s$} \\ 
\sum_{t} p_{rt}(1 - p_{rt}) & \text{if $r = s$} \end{cases}.$$
We therefore have,
\begin{equation*}
\begin{split}
\mathbb{E}[Z] &= \lambda_i^{-1} \bm{u}_i^{\top} \mathbb{E}[(\mathbf{A} - \mathbf{P})^2] \bm{u}_i = 
\lambda_i^{-1} \sum_{s=1}^{n} u_{is}^2 \sum_{t=1}^{n} p_{st}(1 - p_{st}) \\ 
&= \lambda_i^{-1} \sum_{s=1}^{n} u_{is}^2 \sum_{t=1}^{n} X_s^{\top}
\mathbf{I}_{p,q} X_{t} (1 - X_{s}^{\top} \mathbf{I}_{p,q} X_t)
\end{split}
\end{equation*} 
Let $\tilde{\lambda}_i$ and $\bm{v}_i$ be an
eigenvalue/eigenvector pair for the eigenvalue problem
\begin{equation}
\label{eq:def_lambda_tilde}
(\mathbf{X}^{\top} \mathbf{X})^{1/2} \mathbf{I}_{p,q}
(\mathbf{X}^{\top} \mathbf{X})^{1/2} \, \bm{v}= \tilde{\lambda}
\bm{v}.
\end{equation} 
We note that if $\tilde{\lambda}_i$ and $\bm{v}_i$ satisfies
Eq.~\eqref{eq:def_lambda_tilde} then $\tilde{\lambda}_i$ and $\bm{u}_i =
\mathbf{X} (\mathbf{X}^{\top} \mathbf{X})^{-1/2} \bm{v}_i$ 
are an eigenvalue/eigenvector pair for the
eigenvalue problem
$$ \mathbf{X} \mathbf{I}_{p,q} \mathbf{X}^{\top} \bm{u} =
\tilde{\lambda} \bm{u}; \quad \tilde{\lambda} \not = 0.$$
Conversely, if $\tilde{\lambda}_i$ and $\bm{u}_i$ are an eigenvalue/eigenvector
pair for $\mathbf{X} \mathbf{I}_{p,q} \mathbf{X}^{\top}$ then
$\tilde{\lambda}_i$ and 
$\bm{v}_i = (\mathbf{X}^{\top} \mathbf{X})^{-1/2} \mathbf{X}^{\top}
\bm{u}_i$ satisfies Eq.~\eqref{eq:def_lambda_tilde}. 
In addition, if the vectors $\{\bm{v}_i\}_{i=1}^{d}$ are mutually orthonormal
then the vectors $\{\bm{u}_i\}_{i=1}^{d}$ are also mutually orthonormal. We therefore have
\begin{equation*}
\begin{split}
\mathbb{E}[Z] &= 
\tfrac{1}{\lambda_i} \sum_{s=1}^{n} (\bm{v}_i^{\top} (\mathbf{X}^{\top}
\mathbf{X})^{-1/2} X_s)^2 \sum_{t=1}^{n} X_s^{\top} \mathbf{I}_{p,q} X_{t} (1 -
X_{s}^{\top} \mathbf{I}_{p,q} X_t) 
\\ &= 
\tfrac{1}{\lambda_i} \sum_{s=1}^{n} \bm{v}_i^{\top} (\mathbf{X}^{\top}
\mathbf{X})^{-1/2} X_s X_s^{\top} (\mathbf{X}^{\top}
\mathbf{X})^{-1/2} \bm{v}_i X_s^{\top} \mathbf{I}_{p,q}
\bigl(\sum_{t=1}^{n} (X_t - X_t X_t^{\top} \mathbf{I}_{p,q} X_s)
\bigr) 
\\ &= 
\tfrac{n}{\lambda_i} \Bigl(\tfrac{1}{n} \sum_{s=1}^{n} \bm{v}_i^{\top} (\tfrac{\mathbf{X}^{\top}
\mathbf{X}}{n})^{-1/2} X_s X_s^{\top} (\tfrac{\mathbf{X}^{\top}
\mathbf{X}}{n})^{-1/2} \bm{v}_i X_s^{\top} \mathbf{I}_{p,q}
\bigl(\tfrac{1}{n} \sum_{t=1}^{n} (X_t - X_t X_t^{\top} \mathbf{I}_{p,q} X_s) 
\bigr) \Bigr) = \tilde{\eta}_i.
\end{split}
\end{equation*}
By the strong law of large numbers
$$\frac{1}{n} \sum_{t} X_t \rightarrow \mu, \,\,
\frac{\mathbf{X}^{\top} \mathbf{X}}{n} = \frac{1}{n} \sum_{t} X_t X_t^{\top} \rightarrow
\Delta, \,\, 
\frac{\lambda_i}{n} \rightarrow
\lambda_i(\Delta \mathbf{I}_{p,q}).$$ 
as $n \rightarrow \infty$. In addition, when $\lambda_i(\mathbf{I}_{p,q} \Delta)$ is a simple eigenvalue, then we also have $\bm{v}_i \rightarrow \bm{\xi}_i$ as $n \rightarrow \infty$.
We therefore have, when $\lambda_i(\Delta \mathbf{I}_{p,q})$ is a simple eigenvalue, that
\begin{equation} 
\label{eq:mu_eigenvalue_SBM}
\mathbb{E}[Z] \rightarrow \frac{1}{\lambda_i(\Delta \mathbf{I}_{p,q})} \mathbb{E}[\bm{\xi}_i^{\top}
  \Delta^{-1/2} X X^{\top} \Delta^{-1/2} \bm{\xi}_i (X^{\top}
\mathbf{I}_{p,q} \mu - X^{\top} \mathbf{I}_{p,q} \Delta
\mathbf{I}_{p,q} X)]
\end{equation}
as desired. 

{\bf Proof of Eq.~\eqref{eq:u.t(A-P)u_normal}} 
We recall that, conditional on $\mathbf{P}$, $\bm{u}_i^{\top} (\mathbf{A} -
\mathbf{P}) \bm{u}_i$ is a sum of mean zero random
variables. Therefore, by the Lindeberg-Feller central limit theorem
for triangular arrays, we have $\sigma_{i}^{-1} \bm{u}_i^{\top} (\mathbf{A} -
\mathbf{P}) \bm{u}_i$ converges to standard normal; here $\sigma^{2}_{i} = \mathrm{Var}[\bm{u}_i^{\top} (\mathbf{A} - \mathbf{P}) \bm{u}_i]$. All that remains is to evaluate $\sigma^{2}_i$. 
Since $\mathbf{A} - \mathbf{P}$ is symmetric, we have
\begin{equation*}
\begin{split}
\sigma^{2}_i &= \mathrm{Var}\Bigl[\sum_{k < l} (a_{kl} - p_{kl}) (u_{ik} u_{il} + u_{il} u_{ik}) + \sum_{k} (a_{kk} - p_{kk}) u_{ik}^2\Bigr] \\
&= \sum_{k < l} 4 u_{ik}^2 u_{il}^2 p_{kl} (1 - p_{kl}) + \sum_{k} p_{kk}(1 - p_{kk}) u_{ik}^4 
\\ &= 2 \sum_{k} \sum_{l} u_{ik}^2 u_{il}^2 p_{kl} (1 - p_{kl}) - \sum_{k} p_{kk} (1 - p_{kk}) u_{ik}^4 \\
&= 2 \sum_{k} \sum_{l} (X_k^{\top} (\mathbf{X}^{\top} \mathbf{X})^{-1/2}
\bm{v}_i)^2 (X_l^{\top} (\mathbf{X}^{\top} \mathbf{X})^{-1/2} \bm{v}_i)^2 X_k^{\top}
\mathbf{I}_{p,q} X_l (1 - X_k^{\top} \mathbf{I}_{p,q} X_l) + o_{\mathbb{P}}(1) 
\\ &= 2 \Bigl(\sum_{k} (X_k^{\top} (\mathbf{X}^{\top} \mathbf{X})^{-1/2}
\bm{v}_i)^2 X_k^{\top} \Bigr) \mathbf{I}_{p,q} \Bigl(\sum_{l}
(X_l^{\top} (\mathbf{X}^{\top} \mathbf{X})^{-1/2}
\bm{v}_i)^2 \Bigr) + o_{\mathbb{P}}(1) \\ & - 2 \mathrm{tr}
\Bigl(\sum_{k} (X_k^{\top} (\mathbf{X}^{\top} \mathbf{X})^{-1/2}
\bm{v}_i)^2   X_k X_k^{\top} \Bigr) \mathbf{I}_{p,q} \Bigl(\sum_{l}
(X_l^{\top} (\mathbf{X}^{\top} \mathbf{X})^{-1/2} \bm{v}_i)^2 
X_l X_l^{\top} \Bigr) \mathbf{I}_{p,q} 
\end{split}
\end{equation*}
When $\lambda_i(\mathbf{I}_{p,q} \Delta)$ is a simple eigenvalue, the strong law of large numbers and Slutsky's theorem implies, 
\begin{equation}
  \begin{split}
\label{eq:gamma-ii}
\sigma^{2}_i &\rightarrow \bigl(\mathbb{E}[\bm{\xi}_i^{\top} \Delta^{-1/2} X X^{\top}
\Delta^{-1/2} \bm{\xi}_i X]^{\top}
\mathbf{I}_{p,q}
\mathbb{E}[\bm{\xi}_i^{\top} \Delta^{-1/2} X X^{\top}
\Delta^{-1/2} \bm{\xi}_i X]\bigr) \\
& - 
\mathrm{tr} \bigl(\mathbb{E}[
\bm{\xi}_i^{\top} \Delta^{-1/2} X X^{\top}
\Delta^{-1/2} \bm{\xi}_i X X^{\top}]
\mathbf{I}_{p,q}
\mathbb{E}[
\bm{\xi}_i^{\top} \Delta^{-1/2} X X^{\top}
\Delta^{-1/2} \bm{\xi}_i X X^{\top}] \mathbf{I}_{p,q} \bigr) = \Gamma_{ii}
\end{split}
\end{equation}
One last application Slutsky's theorem yield $\tfrac{\lambda_i}{\hat{\lambda}_i}
\bm{u}_i^{\top} (\mathbf{A} - \mathbf{P}) \bm{u}_i
\overset{\mathrm{d}}{\longrightarrow} \mathcal{N}(0, \Gamma_{ii}) $ as
desired.
Finally, we show that if the eigenvalues of $\mathbf{I}_{p,q} \Delta$ are all simple eigenvalues, then $(\hat{\lambda}_i -
\lambda_i)_{i=1}^{d} \longrightarrow
\mathrm{MVN}(\bm{\mu}, \bm{\Gamma})$. More
specifically, for any vector $\bm{s} = (s_1, s_2, \dots, s_d)$ in
$\mathbb{R}^{d}$, we have
\begin{equation*} 
\begin{split}
\sum_{i} s_i (\hat{\lambda}_i - \lambda_i) &= \sum_{i} \frac{s_i \lambda_i}{\hat{\lambda}_i} \bm{u}_i^{\top} (\mathbf{A} -
\mathbf{P}) \bm{u}_i + 
\sum_{i} \frac{s_i \lambda_i}{\hat{\lambda}_i^2} \bm{u}_i^{\top} (\mathbf{A} - \mathbf{P})^{2} \bm{u}_i + o_{\mathbb{P}}(1)
\\ &= \sum_{i} s_i \bm{u}_i^{\top} (\mathbf{A} - \mathbf{P}) \bm{u}_i + \sum_{i} s_i \mu_i + o_{\mathbb{P}}(1)
\\ &= \mathrm{tr} \Bigl((\mathbf{A} - \mathbf{P}) \bigl(\sum_{i} s_i \bm{u}_i \bm{u}_i^{\top}\bigr)\Bigr) + \sum_{i} s_i \mu_i + o_{\mathbb{P}}(1) 
\end{split}
\end{equation*} 
Now let $\mathbf{H} = \sum_{i} s_i \bm{u}_i
\bm{u}_i^{\top}$. Then conditional on $\mathbf{P}$, $\mathrm{tr}
\bigl((\mathbf{A} - \mathbf{P}) \mathbf{H}\bigr)$ is once again a sum
of independent mean $0$ random variables.
Letting $h_{ij}$ be the $ij$-th entry of $\mathbf{H}$, we have
\begin{equation*} 
\begin{split}
\mathrm{Var}\Bigl(\mathrm{tr} \Bigl((\mathbf{A} - \mathbf{P}) \mathbf{H}\Bigr)\Bigr) &= 2 \sum_{k} \sum_{l} p_{kl} (1 - p_{kl}) h_{kl}^2 + o_{\mathbb{P}}(1) \\
&= 2 \sum_{k} \sum_{l} p_{kl} (1 - p_{kl}) (\sum_{i} s_i u_{ik} u_{il})^{2} + o_{\mathbb{P}}(1) \\
&= 2 \sum_{i} \sum_{j} s_i s_j \sum_{k} \sum_{l} p_{kl} (1 - p_{kl}) u_{ik} u_{il} u_{jk} u_{jl}  + o_{\mathbb{P}}(1) \\
\end{split}
\end{equation*} 
which converges to $\sum_{i} \sum_{j} s_{i} s_{j}
\Gamma_{ij}$ as $n \rightarrow \infty$ where $\Gamma_{ij}$ is as
defined in Eq.~\eqref{eq:def_gamma_ij}. Thus for any $\bm{s}$,
$\sum_{i} s_i (\hat{\lambda}_i - \lambda_i) \rightarrow
N(\bm{s}^{\top} \bm{\mu}, \bm{s}^{\top} \bm{\Gamma} \bm{s})$ and hence
by the Cramer-Wold device, $(\hat{\lambda}_i - \lambda_i)_{i=1}^{d} \longrightarrow
\mathrm{MVN}(\bm{\mu}, \bm{\Gamma})$. 

\subsection*{Acknowledgements} We thank Carey E. Priebe and Joshua Cape for illuminating discussions during the writing of this manuscript. 
\bibliography{biblio}

\begin{thebibliography}{32}
\providecommand{\natexlab}[1]{#1}
\providecommand{\url}[1]{\texttt{#1}}
\expandafter\ifx\csname urlstyle\endcsname\relax
  \providecommand{\doi}[1]{doi: #1}\else
  \providecommand{\doi}{doi: \begingroup \urlstyle{rm}\Url}\fi

\bibitem[Airoldi et~al.(2008)Airoldi, Blei, Fienberg, and Xing]{Airoldi2008}
E.~M. Airoldi, D.~M. Blei, S.~E. Fienberg, and E.~P. Xing.
\newblock {Mixed membership stochastic blockmodels}.
\newblock \emph{The Journal of Machine Learning Research}, 9:\penalty0
  1981--2014, 2008.

\bibitem[Arnold(1967)]{Arnold}
L.~Arnold.
\newblock On the asymptotic distribution of the eigenvalues of random matrices.
\newblock \emph{Journal of Mathematical Analysis and Applications},
  20:\penalty0 262–--268, 1967.

\bibitem[Avrachenkov et~al.(2015)Avrachenkov, Cottatellucci, and
  Kadavankandy]{avrachenkov}
K.~Avrachenkov, L.~Cottatellucci, and A.~Kadavankandy.
\newblock Spectral properties of random matrices for stochastic block model.
\newblock In \emph{Proceedings of the $4^{\textrm{th}}$ International Workshop
  on Physics-Inspired Paradigms in Wireless Communications and Networks}, pages
  537--544, 2015.

\bibitem[Benaych-Georges and Nadakuditi(2011)]{benaych-georges}
F.~Benaych-Georges and R.~R. Nadakuditi.
\newblock The eigenvalues and eigenvectors of finite, low rank perturbations of
  large random matrices.
\newblock \emph{Advances in Mathematics}, 227:\penalty0 494--521, 2011.

\bibitem[Bordenave and Capitaine(2016)]{bordenave}
C.~Bordenave and M.~Capitaine.
\newblock Outlier eigenvalues for deformed i.i.d. random matrices.
\newblock \emph{Communications on Pure and Applied Mathematics}, 69:\penalty0
  2131--2194, 2016.

\bibitem[Boucheron et~al.(2003)Boucheron, Lugosi, and Massart]{boucheron2003}
S.~Boucheron, G.~Lugosi, and P.~Massart.
\newblock Concentration inequalities using the entropy method.
\newblock \emph{Annals of Probability}, 31:\penalty0 1583--1614, 2003.

\bibitem[Boucheron et~al.(2013)Boucheron, Lugosi, and
  Massart]{boucheron13:_concen_inequal}
S.~Boucheron, G.~Lugosi, and P.~Massart.
\newblock \emph{Concentration Inequalities: A nonasymptotic theory of
  independence}.
\newblock Oxford University Press, 2013.

\bibitem[Cape et~al.(2016)Cape, Tang, and Priebe]{cape_16_conc}
J.~C. Cape, M.~Tang, and C.~E. Priebe.
\newblock The kato-temple inequality and eigenvalue concentration.
\newblock Arxiv preprint. \url{http://arxiv.org/abs/1603.06100}, 2016.

\bibitem[Davis and Kahan(1970)]{davis70}
C.~Davis and W.~Kahan.
\newblock The rotation of eigenvectors by a pertubation. {III}.
\newblock \emph{Siam Journal on Numerical Analysis}, 7:\penalty0 1--46, 1970.

\bibitem[Ding and Jiang(2010)]{ding10}
X.~Ding and T.~Jiang.
\newblock Spectral distributions of adjacency and laplacian matrices of random
  graphs.
\newblock \emph{Annals of Applied Probability}, 20:\penalty0 2086--2117, 2010.

\bibitem[Erd\H{o}s et~al.(2009)Erd\H{o}s, Schlein, and Yau]{Erdos3}
L.~Erd\H{o}s, B.~Schlein, and H.-T. Yau.
\newblock Local semicircle law and complete delocalization for wigner random
  matrices.
\newblock \emph{Communications in Mathematical Physics}, 287:\penalty0
  641--655, 2009.

\bibitem[Erd\H{o}s et~al.(2010)Erd\H{o}s, P\'{e}ch\'{e}, Ramirez, Schlein, and
  Yau]{Erdos2}
L.~Erd\H{o}s, S.~P\'{e}ch\'{e}, J.~A. Ramirez, B.~Schlein, and H.-T. Yau.
\newblock Bulk universality for {W}igner matrices.
\newblock \emph{Communications on Pure and Applied Mathematics}, 63:\penalty0
  895--925, 2010.

\bibitem[F{\"u}redi and Koml{\'o}s(1981)]{furedi1981eigenvalues}
Z.~F{\"u}redi and J.~Koml{\'o}s.
\newblock The eigenvalues of random symmetric matrices.
\newblock \emph{Combinatorica}, 1:\penalty0 233--241, 1981.

\bibitem[Holland et~al.(1983)Holland, Laskey, and Leinhardt]{holland}
P.~W Holland, K.~B. Laskey, and S.~Leinhardt.
\newblock Stochastic blockmodels: first steps.
\newblock \emph{Social Networks}, 5:\penalty0 109--137, 1983.

\bibitem[Karrer and Newman(2011)]{karrer2011stochastic}
B.~Karrer and M.~E.~J. Newman.
\newblock Stochastic blockmodels and community structure in networks.
\newblock \emph{Physical Review E}, 83:\penalty0 016107, 2011.

\bibitem[Knowles and Yin(2013)]{knowles}
A.~Knowles and J.~Yin.
\newblock Eigenvector distribution of {W}igner matrices.
\newblock \emph{Probability theory and related fields}, 155:\penalty0 543--582,
  2013.

\bibitem[Knowles and Yin(2014)]{knowles_yin}
A.~Knowles and J.~Yin.
\newblock The outliers of a deformed wigner matrix.
\newblock \emph{Annals of Probability}, 42:\penalty0 1980--2031, 2014.

\bibitem[Koltchinskii and Gin\'{e}(2000)]{koltchinskii00:_random}
V.~Koltchinskii and E.~Gin\'{e}.
\newblock Random matrix approximation of spectra of integral operators.
\newblock \emph{Bernoulli}, 6:\penalty0 113--167, 2000.

\bibitem[Lei(2016)]{lei2014}
J.~Lei.
\newblock A goodness-of-fit test for stochastic block models.
\newblock \emph{Annals of Statistics}, 44:\penalty0 401--424, 2016.

\bibitem[O'Rourke and Renfrew(2014)]{O_Rourke}
S.~O'Rourke and D.~Renfrew.
\newblock Low rank perturbation of large elliptic random matrices.
\newblock \emph{Electronic Journal of Probability}, 19:\penalty0 1--65, 2014.

\bibitem[O'Rourke et~al.(2013)O'Rourke, Vu, and Wang]{orourke13:_random}
S.~O'Rourke, V.~Vu, and K.~Wang.
\newblock Random perturbation of low rank matrices: Improving classical bounds.
\newblock Arxiv preprint at \url{http://arxiv.org/abs/1311.2657}, 2013.

\bibitem[P\'{e}ch\'{e}(2006)]{peche}
S.~P\'{e}ch\'{e}.
\newblock The largest eigenvalue of small-rank perturbations of hermitean
  random matrices.
\newblock \emph{Probability Theory and Related Fields}, 134:\penalty0 127--173,
  2006.

\bibitem[Pizzo et~al.(2013)Pizzo, Renfrew, and Soshnikov]{pizzo}
A.~Pizzo, D.~Renfrew, and A.~Soshnikov.
\newblock On finite rank deformation of wigner matrices.
\newblock \emph{Annales de l'Institut Henri Poincar\'{e}, Probabilit\'{e}s et
  Statistiques}, 49:\penalty0 64--94, 2013.

\bibitem[Rubin-Delanchy et~al.(2017)Rubin-Delanchy, Priebe, and
  Tang]{rubin_delanchy_grdpg}
P.~Rubin-Delanchy, C.~E. Priebe, and M.~Tang.
\newblock The generalised random dot product graph.
\newblock Arxiv preprint at \url{http://arxiv.org/abs/1709.01235}, 2017.

\bibitem[Soshnikov(1999)]{Soshnikov}
A.~Soshnikov.
\newblock Universality at the edge of the spectrum in {W}igner random matrices.
\newblock \emph{Communications in Mathematical Physics}, 207:\penalty0
  697--733, 1999.

\bibitem[Tang et~al.(2017)Tang, Athreya, Sussman, Lyzinski, Park, and
  Priebe]{tang14:_semipar}
M.~Tang, A.~Athreya, D.~L. Sussman, V.~Lyzinski, Y.~Park, and C.~E. Priebe.
\newblock A semiparametric two-sample hypothesis testing problem for random dot
  product graphs.
\newblock \emph{Journal of Computational and Graphical Statistics},
  26:\penalty0 344--354, 2017.

\bibitem[Tao and Vu(2010)]{Tao2}
T.~Tao and V.~Vu.
\newblock Random matrices: universality of local eigenvalue statistics up to
  the edge.
\newblock \emph{Communications in Mathematical Physics}, 298:\penalty0
  549--572, 2010.

\bibitem[Tao and Vu(2012)]{tao2012random}
T.~Tao and V.~Vu.
\newblock Random matrices: Universal properties of eigenvectors.
\newblock \emph{Random Matrices: Theory and Applications}, 1, 2012.

\bibitem[Wigner(1955)]{Wigner}
E.~P. Wigner.
\newblock Characteristic vetors of bordered matrices with infinite dimensions.
\newblock \emph{Annals of Mathematics}, 62:\penalty0 548--564, 1955.

\bibitem[Young and Scheinerman(2007)]{young2007random}
S.~Young and E.~Scheinerman.
\newblock Random dot product graph models for social networks.
\newblock In \emph{Proceedings of the 5th international conference on
  algorithms and models for the web-graph}, pages 138--149, 2007.

\bibitem[Yu et~al.(2015)Yu, Wang, and Samworth]{samworth}
Y.~Yu, T.~Wang, and R.~J. Samworth.
\newblock A useful variant of the {D}avis-{K}ahan theorem for statisticians.
\newblock \emph{Biometrika}, 102:\penalty0 315--323, 2015.

\bibitem[Zhang et~al.(2014)Zhang, Nadakuditi, and Newman]{Zhang}
X.~Zhang, R.~R. Nadakuditi, and M.~E. Newman.
\newblock Spectra of random graphs with community structure and arbitrary
  degrees.
\newblock \emph{Physical Review E}, 89, 2014.

\end{thebibliography}

\end{document}